\documentclass{article}

\usepackage{microtype}
\usepackage{graphicx}
\usepackage{booktabs} 
\usepackage{xcolor}
\usepackage{multirow}

\usepackage{hyperref}
\usepackage{amsmath}
\usepackage{mathrsfs}
\usepackage{bm}
\usepackage{amssymb}
\DeclareUnicodeCharacter{2212}{-}
\usepackage{caption}
\usepackage{subcaption}

\usepackage[accepted]{icml2021}

\icmltitlerunning{Uncertainty-aware Generalized Adaptive CycleGAN}

\newcommand{\myparagraph}[1]{\vspace{0pt}\noindent{\bf{#1}}}

\begin{document}

\twocolumn[
\icmltitle{Uncertainty-aware Generalized Adaptive CycleGAN}



\icmlsetsymbol{equal}{*}

\begin{icmlauthorlist}
\icmlauthor{Uddeshya Upadhyay}{tue}
\icmlauthor{Yanbei Chen}{tue}
\icmlauthor{Zeynep Akata}{tue}
\end{icmlauthorlist}

\icmlaffiliation{tue}{Explainable Machine Learning, University of Tuebingen}

\icmlcorrespondingauthor{Uddeshya Upadhyay}{uddeshya.upadhyay@uni-tuebingen.de}

\icmlkeywords{Machine Learning, ICML}

\vskip 0.3in
]



\printAffiliationsAndNotice{}  

\begin{abstract}
Unpaired image-to-image translation refers to learning inter-image-domain mapping in an unsupervised manner. Existing methods often learn deterministic mappings without explicitly modelling the robustness to outliers or predictive uncertainty, leading to performance degradation when encountering unseen out-of-distribution (OOD) patterns at test time. To address this limitation, we propose a novel probabilistic method called Uncertainty-aware Generalized Adaptive Cycle Consistency (UGAC), which models the per-pixel residual by generalized Gaussian distribution, capable of modelling heavy-tailed distributions. 
We compare our model with a wide variety of state-of-the-art methods on two challenging tasks: unpaired image denoising in the natural image and unpaired modality prorogation in medical image domains. Experimental results demonstrate that our model offers superior image generation quality compared to recent methods in terms of quantitative metrics such as signal-to-noise ratio and structural similarity.
Our model also exhibits stronger robustness towards OOD test data.
\end{abstract}

\section{Introduction}
\begin{figure}
    \centering
    \includegraphics[width=0.48\textwidth]{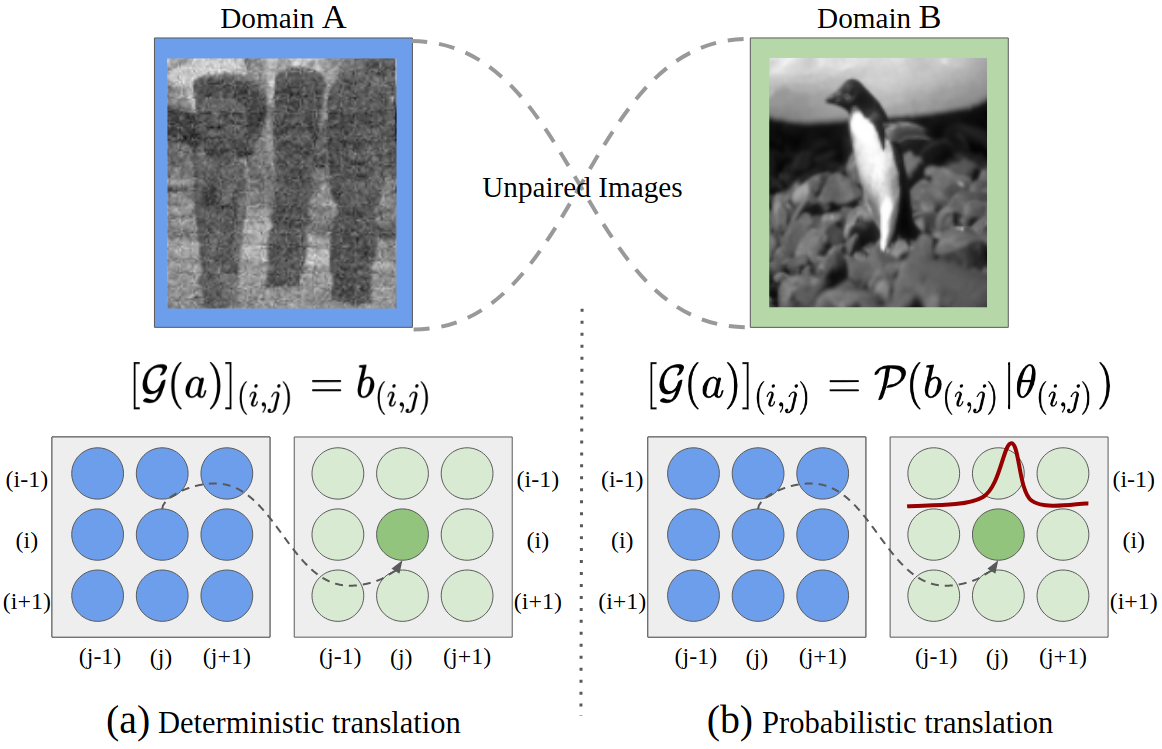}
    \caption{Deterministic vs probabilistic image-to-image translation (between noisy and clean images). (a) Deterministic approach maps a pixel from the input to a pixel from the output domain. (b) Probabilistic approach maps a pixel from the input domain to a probability distribution on the pixel values of the output domain.
    }
    \label{fig:i2i_trans}
\end{figure}
 
Translating an image from a source distribution to an image from a target distribution with a distribution shift (referred to as \textit{domain A} and \textit{domain B} in Figure \ref{fig:i2i_trans}), is an ill-posed problem as a unique deterministic one-to-one mapping may not exist between the two domains. Furthermore, 
since the correspondence between inter-domain samples is missing, their joint-distribution needs to be inferred from a set of marginal distributions. 
However, as infinitely many joint distributions can be decomposed into a fixed set of marginal distributions~\cite{lindvall2002lectures}, the translation problem is ill-posed in the absence of additional constraints. 

Recent deep learning-based methods tackle the unpaired image-to-image translation task by learning inter-domain mappings in a supervised or unsupervised manner. Supervised methods exploit the inter-domain correspondence to learn a mapping between the domains by penalizing the per-pixel residual (using $l_1$ or $l_2$ norm) between the output and corresponding ground-truth sample~\cite{dong2015image,laina2016deeper}.
Unsupervised approaches often train adversarial networks to translate inter-domain samples with an additional constraint on the image space or feature space that imposes structure on the underlying joint distribution of the images from the different domains~\cite{cycleGAN,cut,disGAN}. 

Existing methods often learn a deterministic mapping between the domains. In unsupervised (unpaired) translation from noisy to clean images (see Figure \ref{fig:i2i_trans}-(a)), 
where every pixel in the input domain is mapped to a fixed pixel value in the output domain, 
such a deterministic formulation can lead to mode collapse. Furthermore, a deterministic mapping cannot quantify the model predictive uncertainty to provide trustable model outputs in critical applications, e.g., medical image analysis. 
Finally, existing methods seldom test the model performance on unseen/ out-of-distribution (OOD) perturbed input at test-time, limiting such methods' applicability in the real world. While robustness to outliers is a well-studied problem in statistics~\cite{huber19721972} and optimization~\cite{gentile2003robustness,hastie2015statistical}, it has not attracted as much attention in unpaired translations.

To address these limitations, we propose a new unsupervised probabilistic image-to-image translation method trained without inter-domain correspondence in an end-to-end manner. The probabilistic nature of this method also provides uncertainty estimates for the predictions. Moreover, it is capable of modelling the residuals between the predictions and the ground-truth with heavy-tail distributions, making our model robust to outliers and various unseen data.
Accordingly, we provide a comprehensive analysis on how various state of the art models and our model handle
samples from similar distribution as training-dataset as well as out-of-distribution (OOD) samples, in the context of unsupervised image-to-image translation.

Our contributions are summarized as follows. 
(i) We propose an \textit{unsupervised} probabilistic image-to-image translation model: Uncertainty-aware Generalized Adaptive CycleGAN (UGAC). Our model can model the residuals between the predictions and the ground-truths with heavy-tail distributions that make the model robust to outliers. Probabilistic nature of UGAC also provides uncertainty estimates for the predictions.
(ii) We evaluate UGAC on two tasks: natural image denoising on BSD500 dataset (natural images) and medical image modality propagation on IXI dataset (medical images). We compare our model to seven state-of-the-art image-to-image translation methods. Our results demonstrate that UGAC improves over state of the art, not only on the conventional evaluation setup consisting of in-distribution samples but also on out-of-distribution samples with unseen perturbations and patient-demographics.
%
%
(iii) Lastly, our empirical study shows that our estimated uncertainty scores correlate with the model predictive errors (i.e., residual between model prediction and the ground-truth). This suggests that our uncertainty estimator acts as a good proxy of the model's reliability at test time.
%

\section{Related Work}
Image-to-image translation problems are often formulated as per-pixel deterministic regression between two image domains of~\cite{xie2015holistically,long2015fully,iizuka2016let}. In~\cite{pix2pix}, this is done in a \textit{supervised} (paired) manner using conditional adversarial networks,
while in~\cite{cycleGAN,unit,disGAN,gcGAN,cut} this is done in an \textit{unsupervised} (unpaired) manner by enforcing additional constraints on the joint distribution of the images from separate domains. Both CycleGAN~\cite{cycleGAN} and UNIT~\cite{unit} learn bi-directional mappings, whereas other recent methods~\cite{disGAN,gcGAN,cut} learn uni-directional mappings.

Medical image-to-image translation is of particular interest as it may allow efficiency in diagnosis by synthesizing different modalities algorithmically using only a few imaging modalities~\cite{van2015,yang2020mri,dar2019image,armanious2020medgan}.
However, for such critical applications, confidence in the network's predictions is desirable~\cite{mehta2020uncertainty,seebock2019exploiting}.
Quantification of uncertainty in the predictions made by the unsupervised image-to-image translation models largely remains unexplored. Our proposed method operates at the intersection of uncertainty estimation and unsupervised translation.

Among two broad categories of uncertainties that can be associated with a model's prediction, \textit{epistemic} uncertainty in the model parameters is learned with finite data whereas \textit{aleatoric} uncertainty captures the noise/uncertainty inherent in the data~\cite{kabir2018neural,whatuncer}.
For image-to-image translation, various uncertainties can be estimated using Bayesian deep learning techniques~\cite{whatuncer,probunet,gal2016dropout,lakshminarayanan2017simple,guo2017calibration}. In critical areas like medical imaging, the errors in the predictions deter the adoption of such frameworks in clinical contexts. Uncertainty estimates for the predictions would allow subsequent revision by clinicians~\cite{zhou2020bayesian,hu2019supervised,begoli2019need,ye2020improved,nair2020exploring,wang2019aleatoric,jungo2019assessing}. 

Existing methods model
the per-pixel \textit{heteroscedasticity} as Gaussian distribution for regression tasks~\cite{whatuncer}. This is not optimal in the presence of outliers that often tend to follow heavy-tailed distributions~\cite{oh2016generalized,bouman1993generalized}. Therefore, we enhance the above setup by modelling per-pixel heteroscedasticity as generalized Gaussian distribution, which can model a wide variety of distributions, including Gaussian, Laplace, and heavier-tailed distribution. 

\section{Uncertainty-aware Generalized Adaptive CycleGAN (UGAC) Model } 
\begin{figure}[t]
    \centering
    \includegraphics[width=0.48\textwidth]{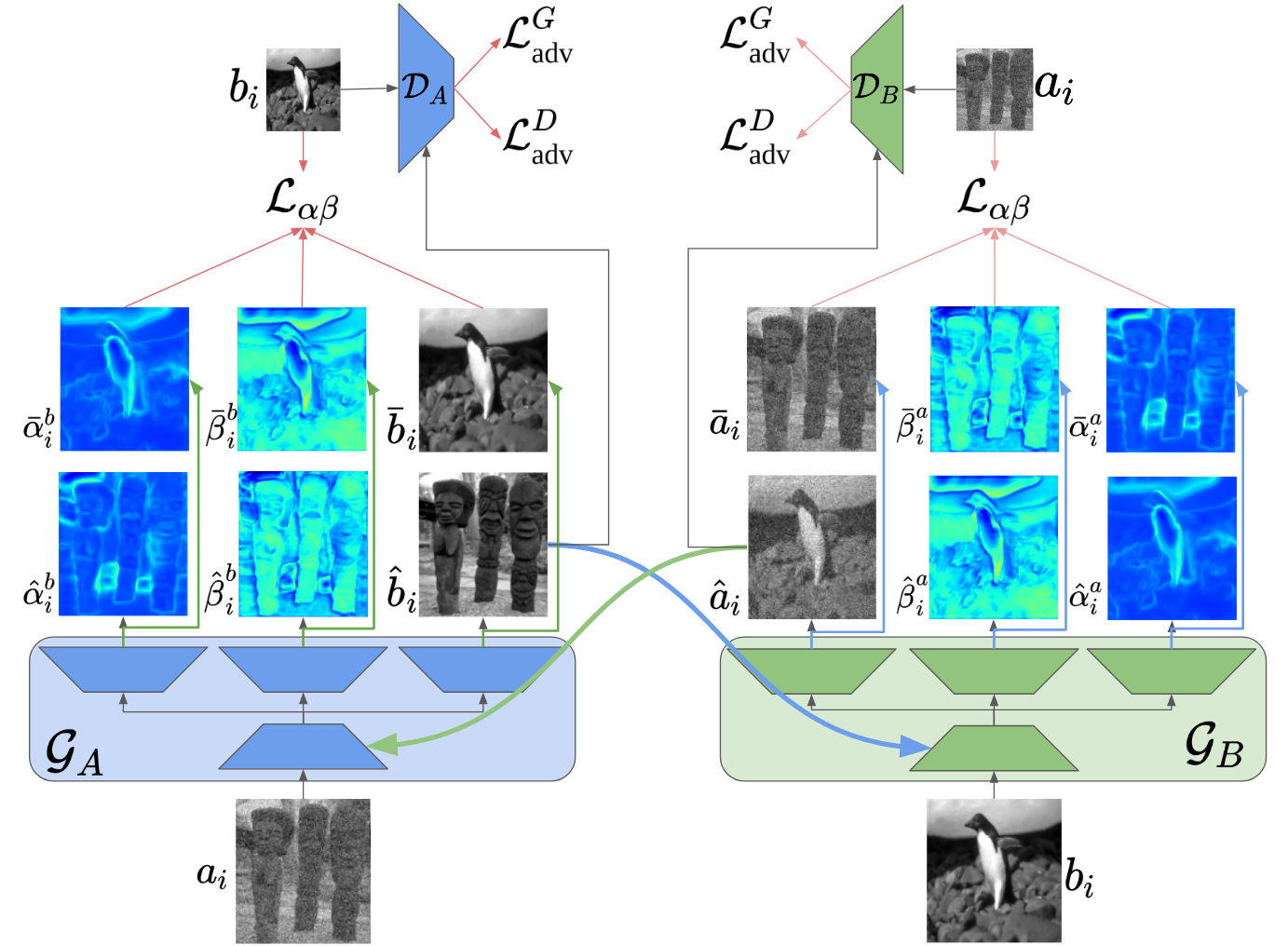}
    \caption{
    Uncertainty-aware Generalized Adaptive CycleGAN showing the cycle between two generators.
    For translating from $A$ to $B$ ($A \rightarrow B$), the input $a_i$ is mapped to generalized Gaussian distribution parameterized by 
    $\{ \hat{b}_i, \hat{\alpha}^b_i, \hat{\beta}^b_i \}$. The backward cycle ($A \rightarrow B \rightarrow A$) reconstructs the image distribution parameterized by
    $\{ \bar{a}_i, \bar{\alpha}^a_i, \bar{\beta}^a_i \}$.
    The method uses
    $\mathcal{L}_{\alpha\beta}$ objective function given by Equation~\ref{eq:ucyc_ab} and adversarial losses given by Equation~\ref{eq:advG} and \ref{eq:advD}. 
    }
    \label{fig:cycle}
\end{figure}

In this section, we present the mathematical formulation of the unsupervised image-to-image translation problem. We discuss the shortcomings of the existing solution involving the cycle consistency loss called CycleGAN~\cite{cycleGAN}. Finally, we present our novel probabilistic framework (UGAC) that overcomes the described shortcomings.

\subsection{Preliminaries} 
\textbf{Formulation.}
Let there be two image domains $A$ and $B$.
Let the set of images from domain $A$ and $B$ be defined by 
(i) $
    S_{A} := \{a_1, a_2 ... a_n\}, \text{ where } a_i \sim \mathcal{P}_A \text{  } \forall  i
$
and
(ii) $
    S_{B} := \{b_1, b_2 ... b_m\}, \text{ where } b_i \sim \mathcal{P}_B \text{  } \forall  i
$, respectively.
The elements $a_i$ and $b_i$ represent the $i^{th}$ image from domain $A$ and $B$ respectively, and are drawn from an underlying \textit{unknown} probability distribution $\mathcal{P}_{A}$ and $\mathcal{P}_{B}$ respectively.

Let each image have $K$ pixels, and $u_{ik}$ represent the $k^{th}$ pixel of a particular image $u_i$.
We are interested in learning a mapping from domain $A$ to $B$ ($A \rightarrow B$) and $B$ to $A$ ($B \rightarrow A$) in an unpaired manner so that the correspondence between the samples from $\mathcal{P}_A$ and $\mathcal{P}_B$ is not required at the learning stage.
In other words, we want to learn the underlying joint distribution $\mathcal{P}_{AB}$ from the given marginal distributions $\mathcal{P}_A$ and $\mathcal{P}_B$. 
This work utilizes CycleGANs that leverage the cycle consistency to learn mappings from both directions ($A \rightarrow B$ and $B \rightarrow A$), but often we are only interested in one direction and the second direction is the auxiliary mapping that aids in learning process.  We define the mapping $A \rightarrow B$ as primary and $B \rightarrow A$ as auxiliary. 

\textbf{Cycle Consistency.}
Learning a joint distribution from the marginal distributions is an ill-posed problem with infinitely many solutions~\cite{lindvall2002lectures}. CycleGAN~\cite{cycleGAN} enforces an additional structure on the joint distribution using a set of primary networks (forming a GAN) and a set of auxiliary networks. The primary networks are represented by $\{\mathcal{G}_A(\cdot; \theta^\mathcal{G}_A), \mathcal{D}_A(\cdot; \theta^\mathcal{D}_A)\}$, where $\mathcal{G}_A$ represents a generator and $\mathcal{D}_A$ represents a discriminator. The auxiliary networks are represented by $\{\mathcal{G}_B(\cdot; \theta^\mathcal{G}_B), \mathcal{D}_B(\cdot; \theta^\mathcal{D}_B)\}$. 
While the primary networks learn the mapping $A \rightarrow B$, the auxiliary networks learn $B \rightarrow A$ (see Figure~\ref{fig:cycle}).

Let the output of the generator $\mathcal{G}_A$ translating samples from domain $A$ (say $a_i$) to domain $B$ be called $\hat{b}_i$. Similarly, for the generator $\mathcal{G}_B$ translating samples from domain $B$ (say $b_i$) to domain $A$ be called $\hat{a}_i$, i.e.,
$
    \hat{b}_i = \mathcal{G}_A(a_i; \theta^{\mathcal{G}}_A) \text{ and } \hat{a}_i = \mathcal{G}_B(b_i; \theta^{\mathcal{G}}_B)
$. 
To simplify the notation, we will omit writing parameters of the networks in the equation.
The cycle consistency constraint~\cite{cycleGAN} re-translates the above predictions ($\hat{b}_i, \hat{a}_i$) to get back the reconstruction in the original domain given by ($\bar{a}_i$,$\bar{b}_i$), where,
\begin{align}
    \bar{a}_i = \mathcal{G}_B(\hat{b}_i) \text{ and }
    \bar{b}_i = \mathcal{G}_A(\hat{a}_i),
\end{align}
and attempts to make reconstructed images ($\bar{a}_i, \bar{b}_i$) similar to original input ($a_i, b_i$) by penalizing the residuals with $\mathcal{L}_1$ norm between the reconstructions and the original input images, giving the cycle consistency ($\mathcal{L}_{\text{cyc}}$),
\begin{align}
    \mathcal{L}_{\text{cyc}}(\bar{a}_i, \bar{b}_i, a_i, b_i) = \mathcal{L}_1(\bar{a}_i, a_i) + \mathcal{L}_1(\bar{b}_i, b_i).
    \label{eq:cyc}
\end{align}

\textbf{Limitations of Cycle consistency.} The underlying assumption when penalizing with the $\mathcal{L}_1$ norm is that the residual at \textit{every pixel} between the reconstruction and the input follow \textit{zero-mean and fixed-variance Laplace} distribution, i.e.,
$\bar{a}_{ij} = a_{ij} + \epsilon^a_{ij}$ and $\bar{b}_{ij} = b_{ij} + \epsilon^b_{ij}$ with,
\begin{align}
    \epsilon^a_{ij}, \epsilon^b_{ij} \sim Laplace(\epsilon; 0,\frac{\sigma}{\sqrt{2}}) \equiv \frac{1}{\sqrt{2\sigma^2}}e^{-\sqrt{2}\frac{|\epsilon-0|}{\sigma}},
\label{eq:lap}
\end{align}
where $\sigma^2$ represents the fixed-variance of the distribution.
This assumption on the residuals between the reconstruction and the input enforces the likelihood (i.e., $\mathscr{L}(\Theta | \mathcal{X}) = \mathcal{P}(\mathcal{X}|\Theta)$, where $\Theta := \theta^{\mathcal{G}}_A \cup \theta^{\mathcal{G}}_B \cup \theta^{\mathcal{D}}_A \cup \theta^{\mathcal{D}}_B$ and $\mathcal{X}:= S_A \cup S_B$) to follow a \textit{factored Laplace} distribution:
\begin{align}
    \mathscr{L}(\Theta | \mathcal{X}) &\propto \bm\prod_{ijpq} e^{-\frac{\sqrt{2}|\bar{a}_{ij}-a_{ij}|}{\sigma}} e^{-\frac{\sqrt{2}|\bar{b}_{pq}-b_{pq}|}{\sigma}},
\end{align}
where minimizing the negative-log-likelihood yields Eq.~\eqref{eq:cyc} with the following limitations.
The residuals in the presence of outliers may not follow the Laplace distribution but instead a heavy-tailed distribution, whereas 
the i.i.d assumption leads to fixed variance distributions for the residuals that do not allow modelling of \textit{heteroscedasticity} to aid in uncertainty estimation. 

\subsection{Building Uncertainty-aware Cycle Consistency}
\label{sec:uncer}
\begin{figure}
    \centering
    \includegraphics[width=0.43\textwidth]{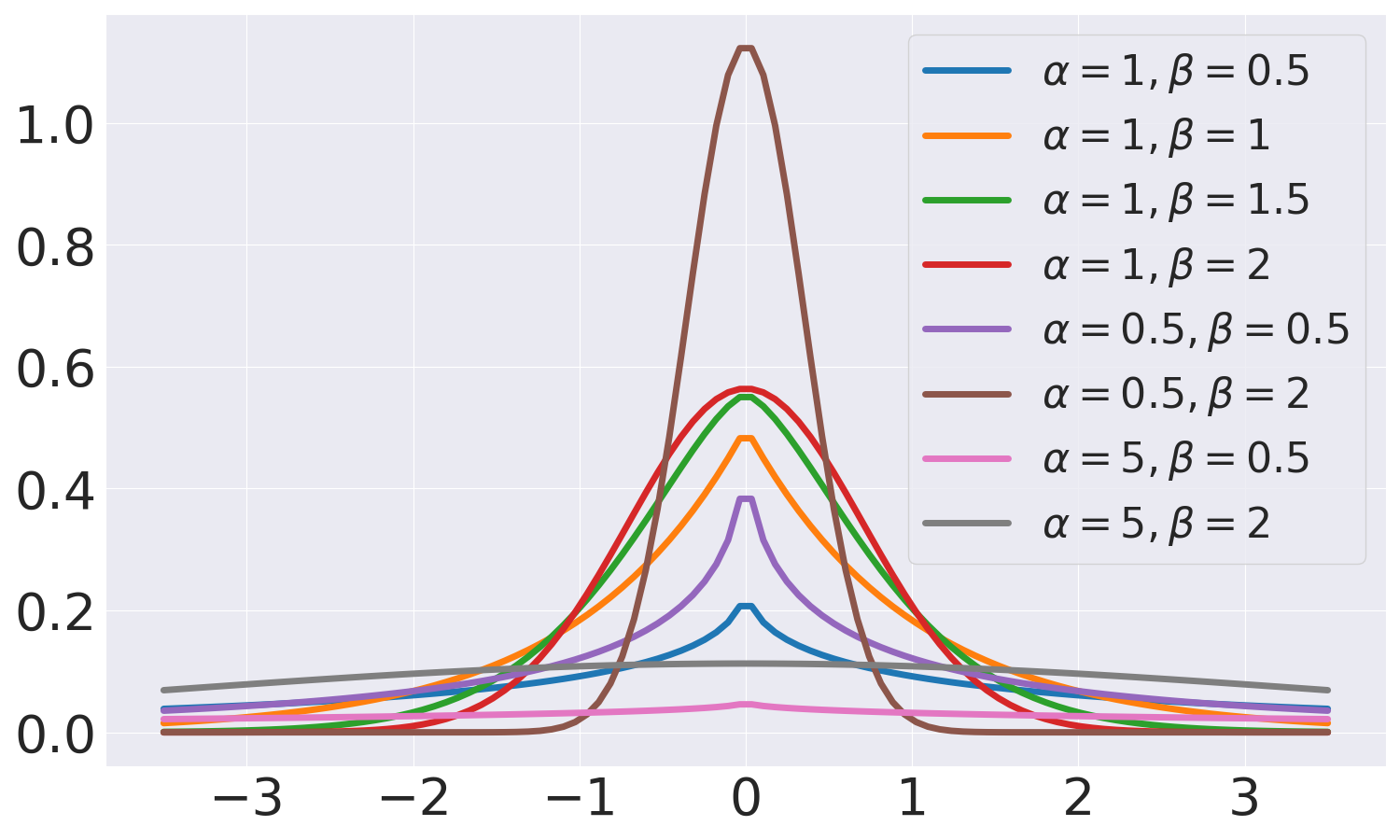}
    \caption{Probability density function (pdf) for generalized Gaussian distribution. Different scale ($\alpha$) and shape ($\beta$) parameters lead to distributions with different tail behaviour. 
    }
    \label{fig:ggd}
\end{figure}

We propose a solution that alleviates the above issues by modelling the underlying per-pixel residual distribution as independent but \textit{non-identically} distributed \textit{zero-mean generalized Gaussian distribution} (GGD) (Figure~\ref{fig:ggd}), i.e., with no fixed shape ($\beta > 0$) and scale ($\alpha > 0$) parameters. Instead, all the shape and scale parameters of the distributions are predicted from the networks and formulated as follows: 
\begin{align}
    \epsilon^a_{ij}, \epsilon^b_{ij} \sim GGD(\epsilon; 0, \bar{\alpha}_{ij}, \bar{\beta}_{ij}) \equiv \frac{\bar{\beta}}{2\bar{\alpha}_{ij}\Gamma(\frac{1}{\bar{\beta}_{ij}})}e^{-\left(\frac{|\epsilon-0|}{\bar{\alpha}_{ij}}\right)^{\bar{\beta}_{ij}}}.
    \label{eq:ggd}
\end{align}
For each $\epsilon_{ij}$, the parameters of the distribution $\{\bar{\alpha}_{ij}, \bar{\beta}_{ij}\}$ may not be the same as parameters for other $\epsilon_{ik}$s; therefore, they are non-identically distributed allowing modelling with heavier tail distributions.
The likelihood for our proposed model is given by,
\begin{equation}
\begin{aligned}
    \mathscr{L}(\Theta | \mathcal{X}) &= 
    \bm\prod_{ijpq}
    \mathscr{G}(\bar{\beta}^a_{ij},\bar{\alpha}^a_{ij},\bar{a}_{ij},a_{ij})
    \mathscr{G}(\bar{\beta}^b_{pq},\bar{\alpha}^b_{pq},\bar{b}_{pq},b_{pq}), 
\end{aligned}
\label{eq:ggd_lhood}
\end{equation}
where ($\bar{\beta}^a_{ij}$) represents the $j^{th}$ pixel of domain $A$'s shape parameter $\beta^a_i$ (similarly for others). $\mathscr{G}(\bar{\beta}^u_{ij},\bar{\alpha}^u_{ij},\bar{u}_{ij},u_{ij})$ is the pixel-likelihood at $j^{th}$ pixel of image $u_i$ (that can represent images of both domain $A$ and $B$) formulated as,
\begin{align}
    \mathscr{G}(\bar{\beta}^u_{ij},\bar{\alpha}^u_{ij},\bar{u}_{ij},u_{ij}) = GGD(u_{ij}; \bar{u}_{ij}, \bar{\alpha}^u_{ij}, \bar{\beta}^u_{ij}),
\end{align}

Hence, minimizing the negative-log-likelihood
yields a new cycle consistency loss, which we call as the uncertainty-aware generalized adaptive cycle consistency loss $\mathcal{L}_{\text{ucyc}}$, given $\mathscr{A}=\{\bar{a}_i, \bar{\alpha}^{a}_i, \bar{\beta}^{a}_i, a_i\}$ and $\mathscr{B}=\{\bar{b}_i, \bar{\alpha}^{b}_i, \bar{\beta}^{b}_i, b_i\}$,
\begin{align}
    \mathcal{L}_{\text{ucyc}}(\mathscr{A}, \mathscr{B}) = 
    \mathcal{L}_{\alpha\beta}(\mathscr{A}) +
    \mathcal{L}_{\alpha\beta}(\mathscr{B}),
    \label{eq:ucyc}
\end{align}
where $\mathcal{L}_{\alpha\beta}(\mathscr{A}) = \mathcal{L}_{\alpha\beta}(\bar{a}_i, \bar{\alpha}^{a}_i, \bar{\beta}^{a}_i, a_i)$ is the new objective function corresponding to domain $A$,
\begin{align}
    & \mathcal{L}_{\alpha\beta}(\bar{a}_i, \bar{\alpha}^{a}_i, \bar{\beta}^{a}_i, a_i) =  \nonumber\\
    & \frac{1}{K}\bm\sum_{j}  \left(\frac{|\bar{a}_{ij}-a_{ij}|}{\bar{\alpha}^{a}_{ij}} \right)^{\bar{\beta}^{a}_{ij}} - 
    \log\frac{\bar{\beta}^{a}_{ij}}{\bar{\alpha}^{a}_{ij}} + \log\Gamma(\frac{1}{\bar{\beta}^{a}_{ij}}),
    \label{eq:ucyc_ab}
\end{align}
where $(\bar{a}_i, \bar{b}_i)$ are the reconstructions for $(a_i,b_i)$ and $(\bar{\alpha}^{a}_i, \bar{\beta}^{a}_i), (\bar{\alpha}^{b}_i, \bar{\beta}^{b}_i) $ are scale and shape parameters for the reconstruction $(\bar{a}_i, \bar{b}_i)$, respectively.

The $\mathcal{L}_1$ norm-based cycle consistency (Eq.~\eqref{eq:cyc}) is a special case of $\mathcal{L}_{\text{ucyc}}$ with 
$(\bar{\alpha}^{a}_{ij}, \bar{\beta}^{a}_{ij}, \bar{\alpha}^{b}_{ij}, \bar{\beta}^{b}_{ij}) = (1,1,1,1) \forall i,j$. 
To utilize $\mathcal{L}_{\text{ucyc}}$, one must have the $\alpha$ maps and the $\beta$ maps for the reconstructions of the inputs. 
To obtain the reconstructed image, $\alpha$ (scale map), and $\beta$ (shape map), we modify the head of the generators (the last few convolutional layers) and split them into three heads, 
connected to a common backbone, as shown in Figure~\ref{fig:cycle}. Therefore, for inputs $a_i$ and $b_i$ to the generator $\mathcal{G}_A$ and $\mathcal{G}_B$, the outputs are:
%
\begin{align}
    (\hat{b}_i, \hat{\alpha}^{b}_i, \hat{\beta}^{b}_i) = \mathcal{G}_A(a_i) \text{ and } 
    (\bar{a}_i, \bar{\alpha}^{a}_i, \bar{\beta}^{a}_i) = \mathcal{G}_B(\hat{b}_i) \nonumber \\
    (\hat{a}_i, \hat{\alpha}^{a}_i, \hat{\beta}^{a}_i) = \mathcal{G}_B(b_i) \text{ and }
    (\bar{b}_i, \bar{\alpha}^{b}_i, \bar{\beta}^{b}_i) = \mathcal{G}_A(\hat{a}_i),
\end{align}

The estimates are plugged into Eq.~\eqref{eq:ucyc} and the networks are trained to estimate all the parameters of the GGD modelling domain $A$ and $B$, i.e. ($\bar{a}_{ij}, \bar{\alpha}^a_{ij}, \bar{\beta}^a_{ij}$) and ($\bar{b}_{ij}, \bar{\alpha}^b_{ij}, \bar{\beta}^b_{ij}$) $\forall ij$.

Furthermore, we apply 
adversarial losses  ~\cite{cycleGAN} to the mapping functions, (i) $\mathcal{G}_A: A \rightarrow B$ and (ii) $\mathcal{G}_B: B \rightarrow A$, using the discriminators $\mathcal{D}_A$ and $\mathcal{D}_B$. The discriminators are inspired from patchGANs~\cite{pix2pix,cycleGAN} that classify whether 70x70 overlapping patches are real or not. The adversarial loss for the generators ($\mathcal{L}_{\text{adv}}^G$~\cite{cycleGAN}) is,
\begin{align}
    \mathcal{L}_{\text{adv}}^G = \mathcal{L}_2(\mathcal{D}^A(\hat{b}_i),1) +
    \mathcal{L}_2(\mathcal{D}^B(\hat{a}_i),1).
    \label{eq:advG}
\end{align}
The loss for discriminators ($\mathcal{L}_{\text{adv}}^D$~\cite{cycleGAN}) is,
\begin{align}
    \mathcal{L}_{\text{adv}}^D = \mathcal{L}_2(\mathcal{D}^A(b_i),1) + \mathcal{L}_2(\mathcal{D}^A(\hat{b}_i),0) + \nonumber \\ 
    \mathcal{L}_2(\mathcal{D}^B(a_i),1) + \mathcal{L}_2(\mathcal{D}^B(\hat{a}_i),0).
    \label{eq:advD}
\end{align}

To train the networks we update the generator and discriminator sequentially at every step ~\cite{cycleGAN,pix2pix,gan_tut}.
The generators and discriminators are trained to minimize $\mathcal{L}^G$ and $\mathcal{L}^D$ as follows:
\begin{align}
    \mathcal{L}^G = \lambda_1 \mathcal{L}_{\text{ucyc}} + \lambda_2 \mathcal{L}_{\text{adv}}^G \text{ and } \mathcal{L}^D = \mathcal{L}_{\text{adv}}^D.
    \label{eq:eqGD}
\end{align}


\myparagraph{Closed-form solution for aleatoric uncertainty.}
Although predicting parameters of the output image distribution allows to sample multiple images for the same input and compute the uncertainty, modelling the distribution as GGD gives us the uncertainty ($\sigma_{\text{aleatoric}}$) without sampling from the distribution as a closed form solution exists and is given by,
$
    \sigma^2_{\text{aleatoric}} = \frac{\alpha^2\Gamma(\frac{3}{\beta})}{\Gamma(\frac{1}{\beta})}
$.
Epistemic uncertainty ($\sigma_{\text{epistemic}}$) is calculated by multiple forward passes with dropouts activated for the same input and computing the variance across the outputs. We define the total uncertainty ($\sigma$) as $\sigma^2 = \sigma^2_{\text{aleatoric}} + \sigma^2_{\text{epistemic}}$.  

\section{Experiments}
In this section, we first describe our experimental setup (i.e., datasets, evaluation metrics and implementation details) in Section \ref{sec:setup}. We compare our model to a wide variety of state-of-the-art methods quantitatively and qualitatively in Section \ref{sec:sota}. Finally, we provide an ablation analysis in Section \ref{sec:ablation} to study the rationale of our model formulation.

\subsection{Experimental Setup}
\label{sec:setup}
\myparagraph{Tasks and Datasets.} 
We use two open-sourced benchmark datasets: (i) BSD500~\cite{amfm_pami2011} for natural image denoising, i.e., mapping noisy images to clean images, without having the ground-truth clean image corresponding to noisy image during training, and (ii) IXI \footnote{from \href{https://brain-development.org/ixi-dataset/}{https://brain-development.org/ixi-dataset/}} for modality propagation in medical imaging. 

BSD500 consists of 500 natural images (300/100/100 for training/val/test).
We convert the images to grayscale and 
following \cite{gnanasambandam2020one}, we evaluate the model on test images with varying levels of noise. 
The noisy images form four noise-levels that are NL0, NL1, NL2, and NL3, where the standard deviation of the Gaussian white noise is set to 0.1, 0.13, 0.17, and 0.20, respectively. This evaluation setup offers the testbed to evaluate the model robustness towards noisy inputs. 

IXI is a medical imaging dataset, including different kinds of MRI scans for $\sim$500 patients (200/100/200 for training/val/test).  We use two common imaging modalities used for diagnosing, T1-weighted MRI (T1w) and T2-weighted MRI (T2w). 
As T1w and T2w MRI from the same patient in the same orientation are often not available and T2w takes longer to acquire, 
learning an unpaired mapping from T1w to T2w is desirable. 
We train the model on clean images. 
However, as the MRI acquisition can often be a noisy process where the noise is Gaussian perturbations in the image space at high SNR~\cite{mri_noise1}, we further evaluate the model on test images with varying levels of noises (referred as NL1, NL2, NL3), where the noises follow Gaussian distribution with standard deviation as 0.05, 0.1, and 0.15 respectively. Similar to the first task, the test images with noises at higher NLs are unseen during training.


\begin{figure}[t]
    \centering
    \includegraphics[width=0.48\textwidth]{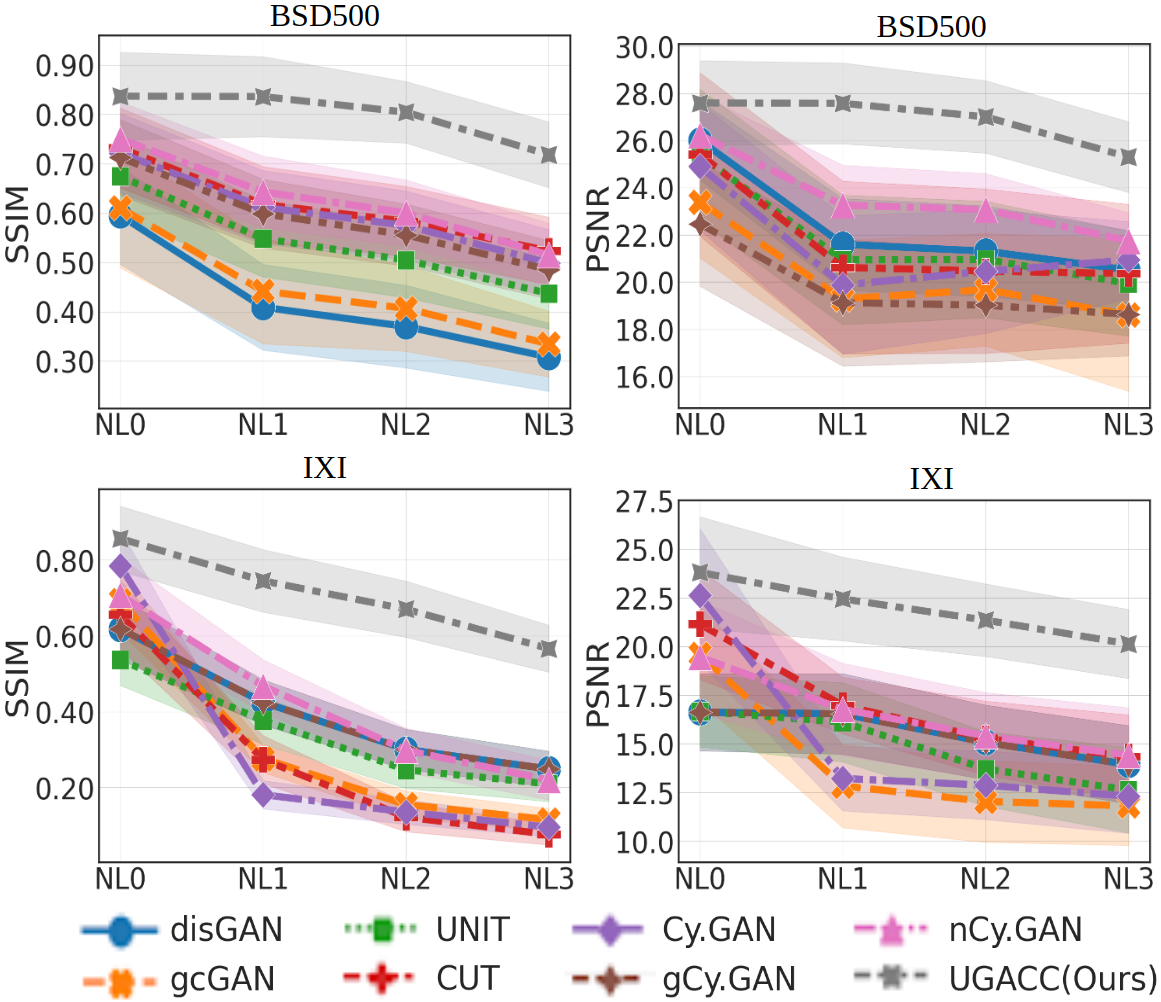}
    \caption{
    Comparing with the state-of-the-art on BSD500 (top) and IXI (bottom) with different noise-levels (NLs). NL0 is the same noise-level as training, NL1, NL2, NL3 are unseen noise-levels (Section~\ref{sec:setup}). 
    We evaluate SSIM (left) and PSNR (right). 
    }
    \label{fig:quant_graphs}
\end{figure}

\begin{figure*}[t]
    \centering
    \includegraphics[width=0.98\textwidth]{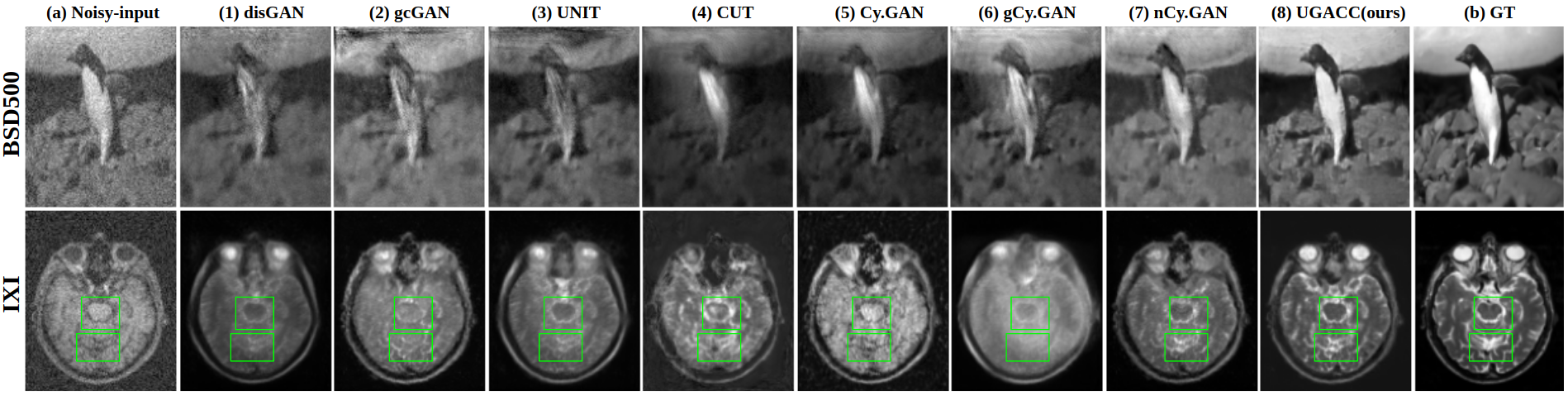}
    \caption{
    Qualitative results for denoising and modality propagation task with BSD500 and IXI, respectively. Output of two different images from test-set at noise-level 3 (NL3) (unseen at training). \textbf{(a)} Noisy-input, \textbf{(1)--(7)} outputs from baselines, and \textbf{(8)} output from UGAC,
    \textbf{(b)} ground-truth images. Output from proposed method are much closer to ground-truth (better in quality than baselines). 
    }
    \label{fig:qual_bsd500}
\end{figure*}


\myparagraph{Evaluation Metrics.} 
We evaluate models 
on two widely adopted metrics: 
PSNR, i.e. $20 \log{\frac{\text{MAX}_I}{\sqrt{\text{MSE}}}}$, where $\text{MAX}_I$ is the highest possible intensity value in the image and $\text{MSE}$ is the mean-squared-error between two images;  
and 
SSIM, measuring the structural similarity between two images~\cite{wang2004image}. 
For both metrics, higher values indicate better model performance. 

\myparagraph{Implementation Details.} 
In our GAN framework, the generator is a cascaded U-Net that progressively improves the intermediate features to yield high-quality output ~\cite{armanious2020medgan}, the discriminator is a patch discriminator~\cite{pix2pix}. 
All the networks were trained using Adam optimizer~\cite{adam} with a mini-batch size of 4. The initial learning rate was set to $2e^{−4}$ and cosine annealing was used to decay the learning rate over 1000 epochs. The hyper-parameters, $(\lambda_1, \lambda_2)$ (Eq.~\eqref{eq:eqGD}) were set to $(10,2)$. For
numerical stability, the proposed network produces $\frac{1}{\alpha}$ instead of $\alpha$. The positivity constraint on the output is enforced by applying the ReLU activation function at the end of the three output layers in the network (Figure~\ref{fig:cycle}). 

\subsection{Comparing with the State of the Art}
\label{sec:sota}

\myparagraph{Compared methods.} 
We compare our model to a wide variety of representative state-of-the-art methods for unsupervised image-to-image translation, including
(1) distanceGAN~\cite{Benaim2017OneSidedUD} (disGAN):  a uni-directional method to map different domains by maintaining a distance metric between samples of the domain with the help of a GAN framework. 
(2) geometry consistent GAN~\cite{gcGAN} 
(gcGAN): a uni-directional method that imposes pairwise distance and geometric constraints.
(3) UNIT~\cite{unit}: a bi-directional method that matches the latent representations of the two domain. 
(4) CUT~\cite{cut}: a uni-directional method that uses contrastive learning to match the patches in the same locations in both domains.
(5) CycleGAN~\cite{cycleGAN} (Cy.GAN): a bi-directional method that uses cycle consistency loss. 
(6) guess Cycle GAN~\cite{advnGAN}: a variant of CycleGAN that uses an additional guess discriminator that ``guesses" at random which of the image is fake in the collection of input and reconstruction images. 
(7) adversarial noise GAN~\cite{advnGAN} (nCy.GAN): another variant of CycleGAN that introduces noise in the cycle consistency loss. 
To ensure a fair comparison, we use the same generator and discriminator architectures for all methods.

\myparagraph{Quantitative Results.} 
We train all models at one noise level (NL0) and evaluate them at varying noise-levels (NL0, NL1, NL2, and NL3) wrt the SSIM and PSNR scores on BSD500 and IXI datasets.

For the unpaired image denoising task on BSD500, in Figure~\ref{fig:quant_graphs} (top), we observe that
our model outperforms the other competitors significantly across all noise-levels. For unseen and higher noise levels (i.e., NL1, NL2, NL3), our model demonstrates a substantial improvement over the seven state of the art methods. 
With increasing the noise level, other methods degrade greatly, while our model does not show significant performance degradation. For instance, the SSIM scores of our model are above 0.70 across all noise levels; while most other approaches obtain SSIM scores lower than 0.65 at the higher noise levels of NL1, NL2 and NL3.
In summary, our results suggest a convincing benefit of our model for unpaired image denoising and demonstrate its robustness to unseen noisy inputs.

For the unpaired modality propagation task on IXI, in Figure~\ref{fig:quant_graphs} (bottom), our model performs significantly better than state of the art under varying noise levels at test time. 
This demonstrates that our model is more robust to noisy input images. As both the generation quality and robustness towards noises are critical in medical imaging, our model indicates a potential practical value for medical imaging analysis applications. 

\myparagraph{Qualitative Results.} 
Figure~\ref{fig:qual_bsd500}-(top) visualizes the generated denoised images from the noisy BSD500 images (at noise level NL3) for all the compared methods. It can be seen that the state of the art methods generate images with artifacts. For instance, we observe blur and visual distortions in the images generated by disGAN, gcGAN, UNIT, gCy.GAN, nCy.GAN (shown in columns 1, 2, 3, 6, and 7, respectively). Furthermore, CUT and CyGAN (shown in columns 4 and 5, respectively) generate dull images that lose the original colour tone where the output images look dark. Similarly, high-frequency features are missing from all the competing baselines (shown in columns 1 to 7).
The result of our model (column 8) demonstrates sharp and clean output images that are closer to the ground-truth.

Figure~\ref{fig:qual_bsd500} (bottom) shows the qualitative results of all the methods for
the task of translating T1w MRI (domain A) to T2w MRI (domain B) in the presence of OOD perturbations (NL3).
Figure~\ref{fig:qual_bsd500}-(bottom (a)) shows the input axial T1w slice and Figure~\ref{fig:qual_bsd500}-(bottom (1) to (8)) show the generated T2w slices. We observe that the other models are unable to properly reconstruct the region around trigeminal-nerve (green bounding box at top) and flax-cerebelli (green bounding box at bottom), which is in sharp contrast to our method that can recover both of these medically relevant structures clearly in the generated T2w images. Moreover, while the high-frequency details throughout the white and grey matter in the brain are missing from the other generated images, the image generated by our model gracefully reconstructs many of the high-frequency details. We present more qualitative results with similar trends in the supplementary.

{
\setlength{\tabcolsep}{2pt}
\renewcommand{\arraystretch}{1.2}
\begin{table*}[t]
\resizebox{\textwidth}{!}{%
\begin{tabular}{llr|cc|cc|cc|cc}
  \multirow{2}{*}{\textbf{Residuals}} & \multirow{2}{*}{\textbf{Loss}} & \multirow{2}{*}{\textbf{Dataset}} & \multicolumn{2}{c}{\textbf{NL0}} & \multicolumn{2}{c}{\textbf{NL1}} & \multicolumn{2}{c}{\textbf{NL2}} & \multicolumn{2}{c}{\textbf{NL3}} \\
  & & & SSIM (std) & PSNR (std) & SSIM (std) & PSNR (std) & SSIM (std) & PSNR (std) & SSIM (std) & PSNR (std)  \\
    \hline
  \multirow{2}{*}{$\frac{1}{\Gamma(\frac{1}{2})}e^{-|\epsilon-0|^2}$ }&
  \multirow{2}{*}{$\left.\mathcal{L}_{\alpha\beta}\right|_{\alpha=1,\beta=2}$} &
  BSD500 & 0.81 (0.03) & 25.7 (1.1) & 0.78 (0.04) & 25.1 (1.1) & 0.72 (0.06) & 24.7 (1.3) & 0.66 (0.05) & 24.1 (1.2)\\ 
  & & IXI & 0.78 (0.07) & 20.7 (2.1) & 0.68 (0.05) & 18.5 (1.9) & 0.60 (0.07) & 17.3 (2.2)  & 0.55 (0.08) & 16.7 (1.8)\\
    \hline
    \multirow{2}{*}{$\frac{1}{2}e^{|\epsilon-0|}$} &
   \multirow{2}{*}{$\left.\mathcal{L}_{\alpha\beta}\right|_{\alpha=1,\beta=1}$} &
   BSD500 & \textbf{0.84 (0.04)} & 26.9 (0.9) & 0.80 (0.02) & 25.5 (1.2) & 0.76 (0.05) & 24.8 (0.8) & 0.69 (0.08) & 24.3 (1.1) \\
  & & IXI & 0.79 (0.06) & 21.6 (2.4) & 0.74 (0.02) & 20.5 (1.8) & 0.66 (0.05) & 19.5 (1.5) & 0.55 (0.07) & 18.1 (1.4) \\
 \hline
  \multirow{2}{*}{$\frac{\beta}{2\alpha\Gamma(\frac{1}{\beta})}e^{-\left (\frac{|\epsilon-0|}{\alpha}\right)^{\beta}}$} &
  \multirow{2}{*}{$\left.\mathcal{L}_{\alpha\beta}\right|_\text{pred}$} &
  BSD500 & 0.83 (0.08) &   \textbf{23.8 (2.9)} & \textbf{0.82 (0.07)} & \textbf{27.2 (1.7)} & \textbf{0.80 (0.06)} & \textbf{26.9 (1.5)} & \textbf{0.72 (0.07)} & \textbf{25.3 (1.3)}\\
  & & IXI & \textbf{0.89 (0.09)} & \textbf{27.5 (1.8)} & \textbf{0.78 (0.08)} &  \textbf{22.5 (2.3)} & \textbf{0.69 (0.09)} & \textbf{21.7 (2.1)} & \textbf{0.58 (0.06)} & \textbf{20.0 (1.7)}
\end{tabular}%
}
\caption{Performance of model with different residual distribution. Table showing the performance across multiple noise-levels for BSD500 and IXI dataset with Laplace, Gaussian and generalized Gaussian distribution modelling the per-pixel residual.}
\label{tab:abl1}
\end{table*}
}

\begin{figure*}[t]
    \centering
    \includegraphics[width=0.98\textwidth]{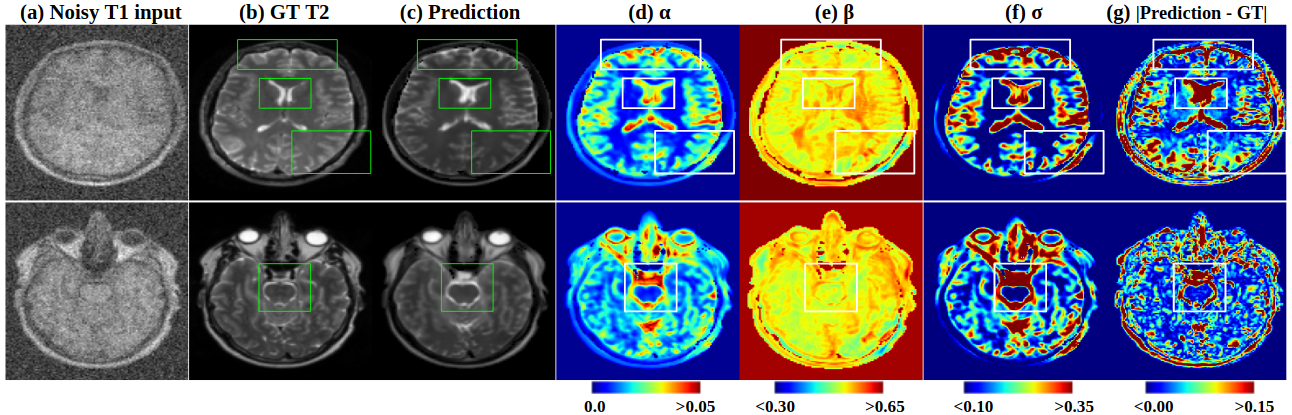}
    \caption{
    Visualization of uncertainty maps for noisy input at NL3.
    \textbf{(a)} Shows the noisy T1w MRI as input.
    \textbf{(b)} Shows the corresponding ground-truth T2w MRI.
    \textbf{(c)} Shows the predicted T2w MRI.
    \textbf{(d)-(e)} Shows the predicted $\alpha$ and $\beta$ maps.
    \textbf{(f)} Shows the uncertainty maps derived from predicted $\alpha$ and $\beta$ maps.
    \textbf{(g)} Shows the absolute residual between the prediction and the ground-truth.
    }
    \label{fig:uncer_viz}
\end{figure*}

\subsection{Analyzing the Model Uncertainty}
\label{sec:ablation}

\myparagraph{Evaluating the generalized adaptive norm.}
We study the performance of our method by modelling the per-pixel residuals in three ways. First, i.i.d Gaussian distribution, i.e., $(\alpha_{ij}, \beta_{ij})$ is manually set to $(1,2) \forall i,j$, which is equivalent to using fixed $l_2$ norm at every pixel in cycle consistency loss $(\left.\mathcal{L}_{\alpha\beta}\right \vert_{\alpha=1, \beta=2})$.
Second,
i.i.d Laplace distribution, i.e., $(\alpha_{ij}, \beta_{ij})$ is manually set to $(1,1) \forall i,j$, which is equivalent to using fixed $l_1$ norm at every pixel in cycle consistency loss $(\left.\mathcal{L}_{\alpha\beta}\right \vert_{\alpha=1, \beta=1})$.
Third, independent but non-identically distributed generalized Gaussian distribution (UGAC), which is equivalent to using spatially varying $l_q$ quasi-norms where $q$ is predicted by the network for every pixel $(\left.\mathcal{L}_{\alpha\beta}\right \vert_{\text{pred}})$.

Table~\ref{tab:abl1} shows the quantitative performance of these three variants across different noise levels for both datasets. We see that spatially adaptive quasi-norms perform better than fixed norms, even at higher noise levels (i.e., in the presence of outliers). 
For instance, at the noise level NL0, our adaptive norm (3rd row) obtains an SSIM score of 0.89 and a PSNR score of 27.5 on IXI dataset. This is significantly higher than the second-best fixed norm formulation, i.e., Laplace distribution with 0.79 and 21.6, respectively. At the highest noise level NL3, our model still holds higher SSIM and PSNR scores (i.e., 0.58 and 20.0) vs. the second-best method with scores of 0.55 and 18.1, respectively. This is consistent with the theory as $l_1$ norm is known to be more robust than $l_2$ norm, and that $l_q$ quasi-norms with ($q < 1$) modelling heavier-tailed distributions are more robust than $l_1$ norm~\cite{gentile2003robustness,oh2016generalized}.


\myparagraph{Visualizing uncertainty maps.} 
We visualize our uncertainty maps for the T1w MRI (domain $A$) to T2w MRI (domain $B$) translation task, on IXI dataset, with OOD perturbations in the input (NL3).
Figure~\ref{fig:uncer_viz}-(a) shows two different input axial slices (T1w at NL3). 
The OOD perturbations have degraded the high-frequency features around the ventricles and trigeminal-nerves in the top and bottom row, respectively (shown within green bounding boxes indicating the region of interests, ROIs).
Figure~\ref{fig:uncer_viz}-(b) shows the corresponding ground-truth axial slices (T2w MRI).

Figure~\ref{fig:uncer_viz}-(c) shows that our method recovers high-frequency details. However, we observe a higher contrast (compared to ground-truth) around the ventricles (top row, green ROI) and trigeminal-nerves (bottom row, green ROI).
The subtle disparity between the contrast in the ventricles, trigeminal-nerves, and grey-matter has been picked up by our scale-map ($\alpha$) and shape-map ($\beta$) as shown in Figure~\ref{fig:uncer_viz}-(d) and (e) respectively.
The pixel-level variation in the $\alpha$ and $\beta$ yields pixel-level uncertainty values in the predictions as described in Section~\ref{sec:setup}. Figure~\ref{fig:uncer_viz}-(f) shows the uncertainty map ($\sigma$) for the predictions made by the network. We see that the disparity in the contrasts between the prediction and the ground-truth is reflected as high uncertainty in the disparity region, i.e., uncertainty is high at ventricles and grey matter region where the reconstruction is of inferior quality (top row, white coloured ROIs). Similarly, uncertainty is high at the trigeminal-nerves (for the bottom row).  

\begin{figure}[t]
    \centering
    \includegraphics[width=0.48\textwidth]{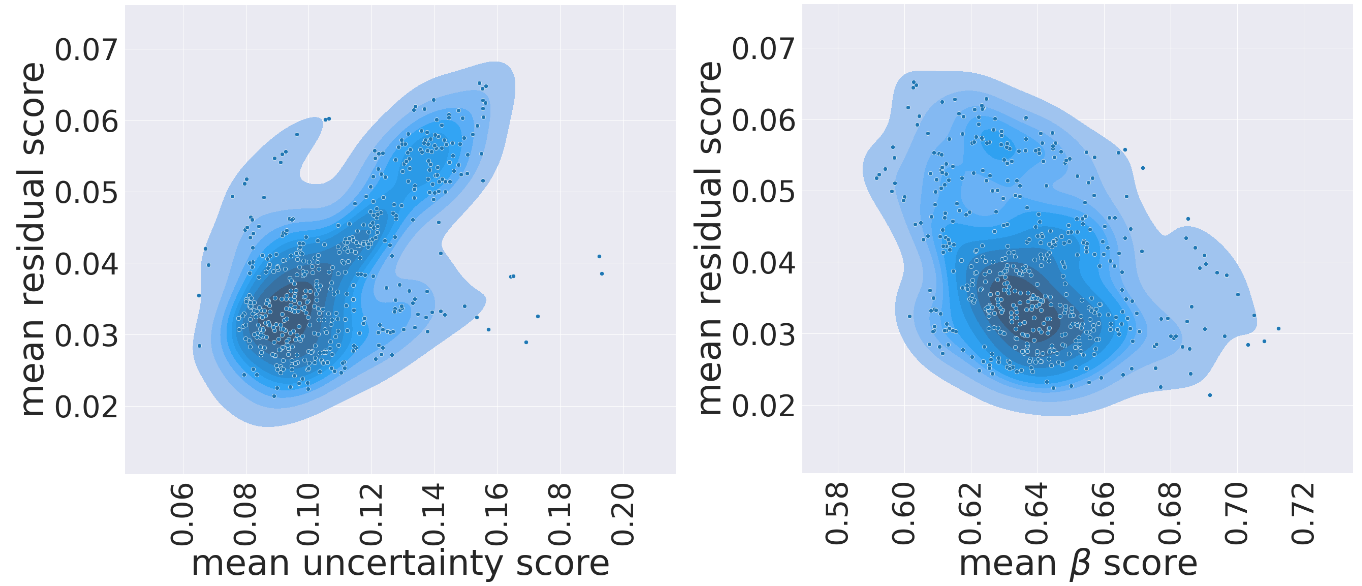}
    \caption{
    Scatterplot between the residual and uncertainty (and $\beta$) values.
    (left) Positive correlation between residual and uncertainty values.
    (right) Negative correlation between residual and $\beta$ values. 
    High-uncertainty corresponds to poor reconstruction. Low $\beta$ values correspond to heavier tailed distributions for residuals. 
    }
    \label{fig:uncer_viz1}
\end{figure}

Our uncertainty maps are in correspondence with the residual maps as shown in Figure~\ref{fig:uncer_viz}-(g), i.e., uncertainty is relatively higher in regions where residual values are higher (indicated by white ROIs in Figure~\ref{fig:uncer_viz}-(g)) and uncertainty values are relatively lower where residual values are low. The correspondence between uncertainty maps (Figure~\ref{fig:uncer_viz}-(f)) and residual maps (Figure~\ref{fig:uncer_viz}-(g)) suggests that uncertainty maps can be used as proxy to residual maps (that are unavailable at the test time, as the ground-truth will not be available) and can serve as an indicator of image quality.
\myparagraph{Uncertainty scores vs. residual scores.} 
To further study the relationship between the uncertainty maps and the residual maps across a wide variety of images in the test-set,
we show the density and the scatter-plot between the residual score and uncertainty score in Figure~\ref{fig:uncer_viz1} (left), where every point represents a single image. 
For an image, the mean residual score (on the $y$-axis) is derived as the mean of absolute residual values for all the pixels in the image. Similarly, the uncertainty score (on the $x$-axis) is calculated as the mean of uncertainty values of all the pixels in that image.
From the plot, we see that across the test-set mean uncertainty score correlates positively with the mean residual score, i.e., higher uncertainty score corresponds to higher residual. An image with higher residual score represents a poor quality image. This further supports the idea that uncertainty maps derived from our method can be used as a proxy to residual that indicates the overall image quality of the output generated by our network.

Figure~\ref{fig:uncer_viz1} (right) shows 
a negative correlation between the mean residual scores and the mean shape-parameter ($\beta$) scores (obtained in a fashion similar to mean uncertainty scores), i.e., lower mean shape-parameter scores corresponds to higher residual scores.
This indicates that for relatively poor quality output (with high residual score), our method models the image with heavier-tailed distribution (lower $\beta$ values corresponds to heavier-tailed distributions, as can be seen in Figure~\ref{fig:ggd}). This is in line with the theory as outliers lead to high residuals and tend to follow heavier-tailed distributions~\cite{gentile2003robustness}.

{
\setlength{\tabcolsep}{3pt}
\renewcommand{\arraystretch}{1.1}
\begin{table}
\centering
\resizebox{0.48\textwidth}{!}{%
\begin{tabular}{l|cccc}
 & \multicolumn{2}{c}{\textbf{OOD type: Age}} & \multicolumn{2}{c}{\textbf{OOD type: Weight}} \\
& \multicolumn{2}{c}{\textbf{young $\rightarrow$ old} } & \multicolumn{2}{c}{\textbf{light $\rightarrow$ heavy}} \\
\textbf{Methods} & SSIM (std) & PSNR (std) & SSIM (std) & PSNR (std) \\ \hline
(1)      & 0.62 (0.05) & 17.1 (0.9) & 0.66 (0.06) & 17.8 (0.9) \\ 
(2)       & 0.67 (0.03) & 18.3 (1.1) & 0.69 (0.07) & 18.9 (0.7) \\ 
(3)       & 0.71 (0.06) & 19.4 (1.4) & 0.74 (0.04) & 19.7 (1.4) \\ 
(4)        & 0.68 (0.08) & 18.6 (1.7) & 0.70 (0.08) & 19.1 (1.5) \\ 
(5)      & 0.70 (0.06) & 19.9 (2.1) & 0.72 (0.09) & 20.6 (1.7) \\ 
(6)     & 0.61 (0.04) & 16.6 (0.8) & 0.64 (0.11) & 18.2 (1.1) \\ 
(7)    & 0.69 (0.09) & 19.2 (1.8) & 0.71 (0.08) & 20.9 (1.9) \\ \hline
UGAC(ours) & \textbf{0.83 (0.07)} & \textbf{25.8 (2.3)} & \textbf{0.85 (0.15)} & \textbf{26.3 (2.1)} \\ 
\end{tabular}%
}
\caption{Performance on OOD data. Table comparing different methods on patient demography related OOD data. (1) disGAN~\cite{disGAN}, (2) gcGAN~\cite{gcGAN}, (3) UNIT~\cite{unit}, (4) CUT~\cite{cut}, (5) Cy.GAN~\cite{cycleGAN}, (6) gCy.GAN~\cite{advnGAN}, (7) nCy.GAN~\cite{advnGAN} }
\label{tab:abl2}
\end{table}
}

\myparagraph{Generalizing to out-of-distribution data.}
To analyze how different models can handle OOD data, we design an experiment based on patient demographics.
We use the metadata from the IXI dataset to split the patient cohort into two non-overlapping groups based on \textit{age} and \textit{weight}. One group is used for training the models while the other for testing. The first setup contains training data of people with age${<}45$ and test data of people with age${\geqslant}45$. The second setup contains training data of people with weight ${<}68$kg and test data of people with weight ${\geqslant}68$kg.

Table~\ref{tab:abl2} shows the performance of all the models when trained and evaluated on different splits. Zooming into the \textit{young} $\rightarrow$ \textit{old} experiments, where the model is trained on ``young'' patients and evaluated on ``old'' patients, our model has an SSIM and PSNR of $0.83$ and $25.8$, respectively, compared to that of $0.71$ and $19.4$ for the best performing baseline. We see a similar trend for the other split training on lighter patients and evaluating on heavier patients (see Table~\ref{tab:abl2}-Col 3). 
These results indicate that our model can generalize better to unseen OOD data. This is because, while the other methods mostly learn deterministic inter-domain mappings, our model introduces a probabilistic formulation to model OOD behaviors explicitly, thus being more robust to translate unseen images at test time.

\section{Conclusion}
In this work, we propose a new uncertainty-aware generalized cycle consistency for unsupervised image-to-image translation along with uncertainty estimation. We demonstrate the efficacy of the proposed method on robust unpaired image denoising (BSD500 dataset) and on robust MRI scan translation (IXI dataset) that often suffers from noise corruptions in the real world. In addition, we show that the uncertainty estimates are faithful to the residuals between the predictions and the ground-truth. We also perform experiments on various kinds of OOD data, including noise-perturbation for natural and medical images and shift in patient demography for medical data, and show that our method outperforms all the baselines by generating superior images in terms of quantitative metrics and appearance.

\bibliography{paper}
\bibliographystyle{icml2021}

\newpage
\appendix
\section{More Qualitative Results}
In this section, we present more qualitative results for all the baselines that were used in the study presented in main manuscript along with our proposed model on both the datasets, i.e., BSD500 consisting of natural images for denoising task and IXI consisting of MRI dataset for modality propagation task. 

We use the following baselines in the study,
(1) distanceGAN (disGAN):  a uni-directional method to map different domains by maintaining a distance metric between samples of the domain with the help of a GAN framework. 
(2) geometry consistent GAN 
(gcGAN): a uni-directional method that imposes pairwise distance and geometric constraints.
(3) UNIT: a bi-directional method that matches the latent representations of the two domain. 
(4) CUT: a uni-directional method that uses contrastive learning to match the patches in the same locations in both domains.
(5) CycleGAN (Cy.GAN): a bi-directional method that uses cycle consistency loss. 
(6) guess Cycle GAN: a variant of CycleGAN that uses an additional guess discriminator that ``guesses" at random which of the image is fake in the collection of input and reconstruction images. 
(7) adversarial noise GAN (nCy.GAN): another variant of CycleGAN that introduces noise in the cycle consistency loss. 
To ensure a fair comparison, we use the same generator and discriminator architectures for all methods.

\begin{figure*}[t]
    \centering
    \includegraphics[angle=90,scale=0.5,width=0.9\textwidth]{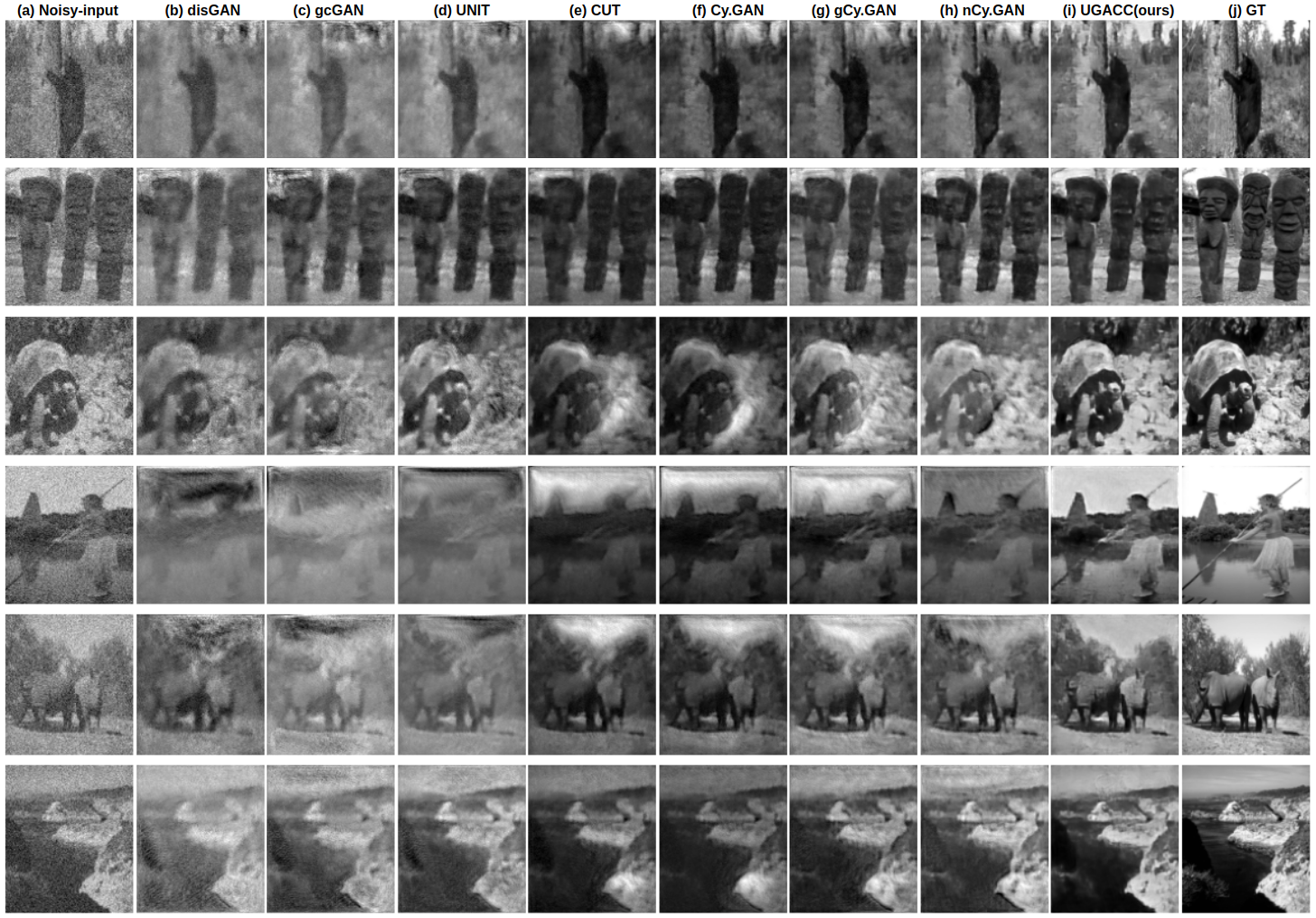}
    \caption{Qualitative results for denoising task with BSD500. Output of different images from test-set at noise-level 3 (NL3) (unseen at training). \textbf{(a)} Noisy-input, \textbf{(b)--(h)} outputs from baselines, and \textbf{(i)} output from UGAC,
    \textbf{(j)} ground-truth images.}
    \label{fig:bsd}
\end{figure*}

\begin{figure*}[t]
    \centering
    \includegraphics[angle=90,scale=0.5,width=0.85\textwidth]{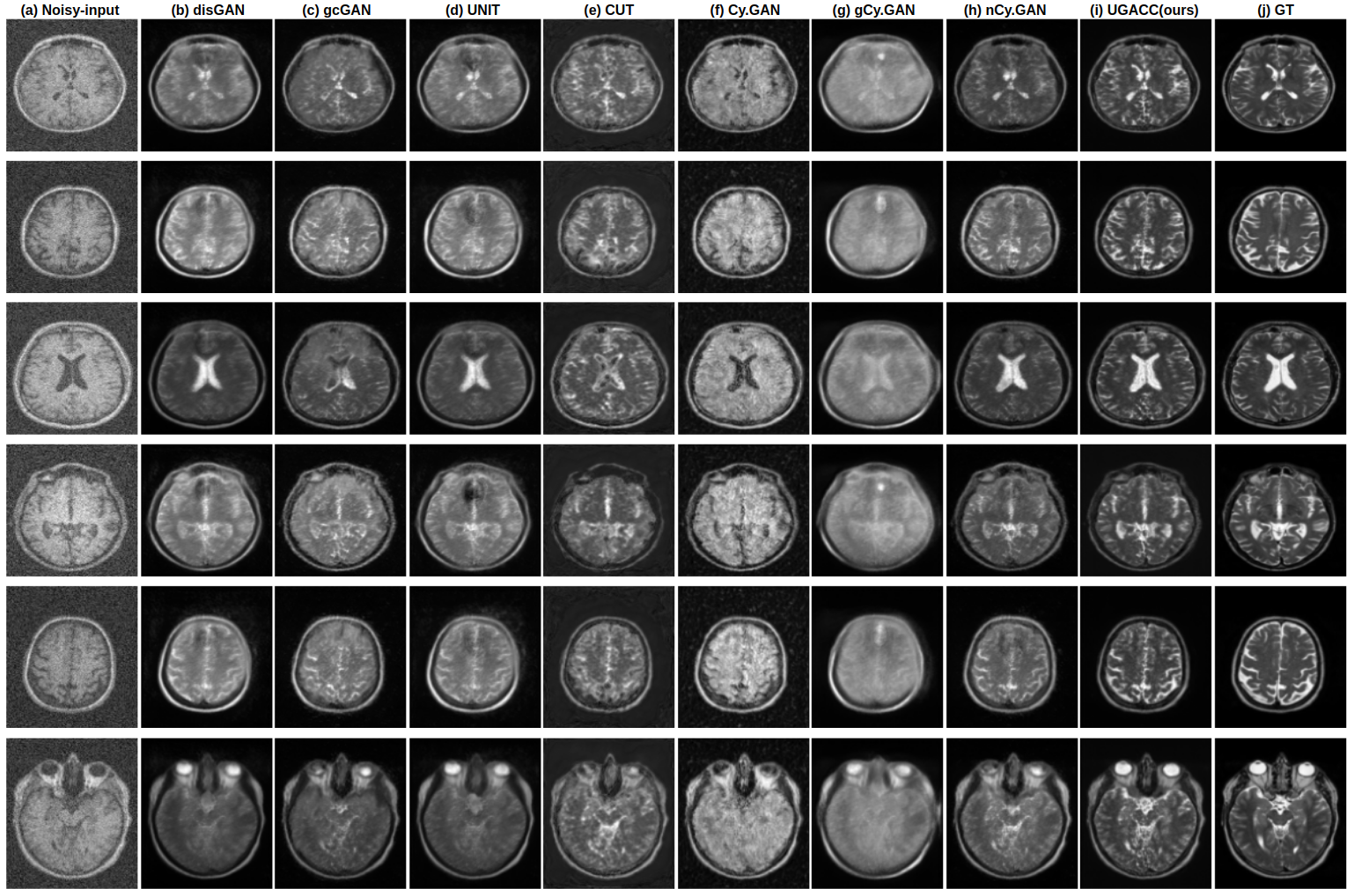}
    \caption{Qualitative results for modality propagation task with IXI. Output of different images from test-set at noise-level 3 (NL3) (unseen at training). \textbf{(a)} Noisy-input, \textbf{(b)--(h)} outputs from baselines, and \textbf{(i)} output from UGAC,
    \textbf{(j)} ground-truth images.}
    \label{fig:ixi}
\end{figure*}

Figure~\ref{fig:bsd} and Figure~\ref{fig:ixi} visualize the (i) generated denoised images from the noisy BSD500 images (at noise level NL3) and (ii) generated T2w images from the T1w images for IXI dataset (at noise level NL3), respectively, for all the methods. It can be seen that the existing methods generate images with artifacts. High-frequency features are missing from all the competing baselines (shown in columns 1 to 7).
The result of our model (column 8) demonstrates sharp and clean output images that are closer to the ground-truth.

\end{document}